\documentclass[10pt,twocolumn,letterpaper]{article}
\PassOptionsToPackage{table}{xcolor}
\usepackage{wacv}

\usepackage{graphicx}
\usepackage{amsmath,amssymb}
\usepackage{array}
\usepackage{booktabs}
\usepackage{tabularx}
\usepackage{xurl}
\usepackage[accsupp]{axessibility}
\definecolor{wacvblue}{rgb}{0.21,0.49,0.74}
\usepackage[pagebackref,breaklinks,colorlinks,allcolors=wacvblue]{hyperref}

\graphicspath{{./}{paper_assets/}}

\def\wacvPaperID{1699}
\def\confName{WACV}
\def\confYear{2027}

\newcommand{\AAS}{AAS}
\newcolumntype{Y}{>{\raggedright\arraybackslash}X}

\title{Rethinking Perceptual Similarity for Remote Sensing: Multi-Layer Attention Scoring and Human Preference Benchmarking}

\author{
Eli Underwood$^{1,2}$ \quad Marc Bosch$^{1}$ \quad Neil Joshi$^{1}$ \quad Jeffrey Wen$^{1}$ \quad Myron Brown$^{1}$\\[2pt]
$^{1}$Johns Hopkins University Applied Physics Laboratory \quad $^{2}$Duke University\\[2pt]
{\tt\small eli.underwood@duke.edu, \{marc.bosch, neil.joshi, jeffrey.wen, myron.brown\}@jhuapl.edu}\\[4pt]
{\small \url{https://github.com/pubgeo/rspd}}
}

\begin{document}

\maketitle

\begin{abstract}
Perceptual similarity metrics for natural images transfer poorly to remote sensing (RS), where severe geospatial and semantic failures can remain visually plausible while still being decisively wrong for downstream use.
To address this, we develop a novel human-aligned perceptual quality metric for generated RS imagery and evaluate it across several generation tasks.
We first conduct a controlled perturbation study where we progressively increase the severity of common synthetic perturbations.
Using the layerwise Aligned Attention Score (\AAS) responses in RemoteCLIP, we find that different RS artifacts are best captured at different depths.
Based on these findings, we introduce Remote Sensing Perceptual Distance (RSPD), a pairwise metric that averages seven independently calibrated layer distances without tuning on human-preference labels.
To evaluate our metric on realistic RS generation tasks, we construct a blind, reference-conditioned two-alternative forced-choice (2AFC) benchmark with five 300-triplet subsets spanning synthetic artifacts, airborne and spaceborne super-resolution, SAR-to-optical translation, and timeseries controls.
We also construct a 300-triplet ordered misregistration benchmark with deterministic labels.
Using the same RemoteCLIP checkpoint, RSPD substantially outperforms simple cosine distance with the final global embedding and is competitive with the strongest generic baselines across the realistic RS generation tasks.
\end{abstract}
\maketitle

\begingroup
\renewcommand{\thefootnote}{}
\footnotetext{This work was supported by the Intelligence Advanced Research Projects Activity (IARPA) contract number 2020-20081800401. Disclaimer: The views and conclusions contained herein are those of the authors and should not be interpreted as necessarily representing the official policies or endorsements, either expressed or implied, of IARPA or the U.S. Government.}
\endgroup

\section{Introduction}
\label{sec:intro}

Remote sensing is no longer only a recognition problem, as foundation models and diffusion models are now used for super-resolution (SR), harmonization, cross-modality translation, style transfer, and semantic editing of overhead imagery~\cite{khanna2024diffusionsatgenerativefoundationmodel,liu2025text2earthunlockingtextdrivenremote,yu2024metaearthgenerativefoundationmodel,sastry2024geosynthcontextuallyawarehighresolutionsatellite,xiao2023ediffsrefficientdiffusionprobabilistic,wang2024semanticguidedlargescale,zhuang2026harmonidiffrstrainingfreediffusionharmonization,zhenyuan2026rsedittextguidedimageediting,han2025exploringtextguidedsingleimage,karatzoglidi2020neuralstyletransferremote,Zhang_2022,Vinholi_2025,do2025cdiffsetleveraginglatentdiffusion,chen2026any2anyunifiedarbitrarymodality,Qi_2026}. In these settings, outputs can look realistic while still being misregistered, oversmoothed, radiometrically distorted, or semantically incorrect. A generated image can look realistic while shifting roads or changing building footprints, so evaluating fidelity to the reference matters for applications such as mapping and change detection. Classical full-reference measures such as SSIM~\cite{wang2004imagequality}, learned perceptual metrics such as LPIPS~\cite{zhang2018unreasonableeffectivenessdeepfeatures} and DreamSim~\cite{fu2023dreamsim}, and newer attention-aware methods such as DiffSim~\cite{song2024diffsimtamingdiffusionmodels} are therefore an imperfect fit for the remote-sensing regime~\cite{Blau_2018}.

RS-specific evaluation frameworks have begun to close this gap. OpenSR-test~\cite{aybar2024opensrtest} decomposes SR quality into consistency, synthesis, and correctness terms, evaluated on cross-sensor datasets curated to minimize spatial and spectral misalignment. Isolating alignment as its own consistency term is what makes the correctness terms interpretable, but it also assumes that candidate and reference are essentially co-registered. That assumption is reasonable for laboratory-style SR evaluation, where the pairing is constructed to satisfy it, and does not extend to cross-modality translation, harmonization, or semantic editing, where the candidate is not pixel-aligned with the reference and misregistration is itself one of the failures to be judged rather than a nuisance removed before judging.

The mismatch has also been shown empirically within RS. In a large crowdsourced study of super-resolution outputs, Wolters et al.~\cite{wolters2023zooming} found PSNR, cPSNR, and SSIM agree with human preference at roughly chance, while cosine similarity between pooled CLIP embeddings agrees with annotators around $85\%$ of the time. We build on both lessons: RS perceptual evaluation needs human data rather than pixel fidelity, and domain-relevant learned representations are the right substrate for it. Two questions remain open. First, \emph{how} those representations should be compared: a pooled embedding per image is one comparison operator among many and may discard spatial detail that matters for RS failures. Second, \emph{over what range of tasks} such a metric should be validated, since conclusions drawn from super-resolution alone may not hold for translation or semantic editing, where failure modes differ in kind rather than degree.

Our working hypothesis is that RS perceptual scoring should be built from domain-aligned representations and stress-tested on RS-specific failure modes before being trusted on real generation tasks. Controlled synthetic perturbations first reveal which RemoteCLIP~\cite{liu2024remoteclipvisionlanguagefoundation} layers are sensitive to which classes of RS artifact. RSPD then averages seven independently calibrated layer distances to give each reference-candidate pair one fixed score without learning weights from preference labels. Human-preference triplets test whether those insights transfer to airborne and spaceborne super-resolution, SAR-to-optical translation, and curated timeseries imagery.

We contribute (1) a controlled artifact-ladder analysis showing that, with the backbone held fixed, intermediate-layer AAS responds more clearly than final-embedding cosine to several RS artifacts, (2) RSPD, a pairwise seven-layer score calibrated only with the controlled benchmark, and (3) an RS human-preference benchmark spanning several generation tasks under one blind annotation protocol (1{,}500 triplets over synthetic artifacts, airborne and spaceborne super-resolution, SAR-to-optical translation, and timeseries controls), plus a 300-triplet ordered misregistration set whose labels follow deterministically from severity ordering. The closest RS preference collection we know of is single-task and pairwise, and served as an internal instrument for choosing among candidate metrics~\cite{wolters2023zooming}.

\section{Related Work}
\label{sec:related}

Perceptual similarity metrics replace pixel fidelity with distances in learned feature spaces, with SSIM and MS-SSIM still widely used as stable structural baselines~\cite{wang2004imagequality,wang2003multiscale}. LPIPS showed pretrained features correlate surprisingly well with human judgments~\cite{zhang2018unreasonableeffectivenessdeepfeatures}, DreamSim used synthetic supervision to capture richer visual relationships~\cite{fu2023dreamsim}, and DiffSim compared attention behavior rather than only pooled embeddings~\cite{song2024diffsimtamingdiffusionmodels}. Human-preference datasets such as PieAPP highlighted the importance of pairwise judgments~\cite{prashnani2018pieapp}, and the perception-distortion tradeoff explains why visually better outputs need not optimize fidelity metrics~\cite{Blau_2018}. These issues are amplified in remote sensing, where perceptual quality must reflect geospatial alignment, object layout, boundaries, and semantic correctness beyond low-level appearance, and where generative models have rapidly expanded the space of outputs requiring evaluation~\cite{khanna2024diffusionsatgenerativefoundationmodel,liu2025text2earthunlockingtextdrivenremote,yu2024metaearthgenerativefoundationmodel,sastry2024geosynthcontextuallyawarehighresolutionsatellite,xiao2023ediffsrefficientdiffusionprobabilistic,wang2024semanticguidedlargescale,zhuang2026harmonidiffrstrainingfreediffusionharmonization,zhenyuan2026rsedittextguidedimageediting,han2025exploringtextguidedsingleimage,karatzoglidi2020neuralstyletransferremote,Zhang_2022,Vinholi_2025,do2025cdiffsetleveraginglatentdiffusion,chen2026any2anyunifiedarbitrarymodality}. RS quality assessment itself remains fragmented, spanning operational image-quality inspection, task-specific SR quality metrics, and dedicated SR benchmarking frameworks~\cite{Yu_2023,cheng2023newsuperresolutionmeasurementperceptual,lin2024perceptionfidelityawarereducedreferencesuperresolution,aybar2024opensrtest}. These target instrument-referenced or task-specific quality on aligned SR pairs rather than blind visual human preference across several RS generation tasks.

\paragraph{Human-aligned metrics for RS generation.} Closest to our work is the SR study of Wolters et al.~\cite{wolters2023zooming}, which collected 11{,}524 pairwise crowdsourced judgments on S2-NAIP and WorldStrat, found PSNR, cPSNR, and SSIM near chance, and adopted cosine similarity between pooled CLIP embeddings as an RS metric and auxiliary loss. We differ in three ways. Primarily, their score compares two final pooled embeddings, whereas we compare intermediate attention at multiple depths. Section~\ref{sec:aas-vs-cosine} uses the same RemoteCLIP checkpoint in both cases and finds intermediate-layer AAS with multi-layer aggregation more effective than final-embedding cosine. Second, their study covers super-resolution alone, whereas our triplets add SAR-to-optical translation, mixed synthetic artifacts, misregistration, and timeseries controls under one blind protocol. Finally, their annotations were an internal instrument for metric selection, whereas our triplet sets are designed from the outset as a reusable evaluation artifact with balanced per-task subsets and a fixed no-tuning protocol.

\begin{figure*}[t]
\centering
\includegraphics[width=\textwidth]{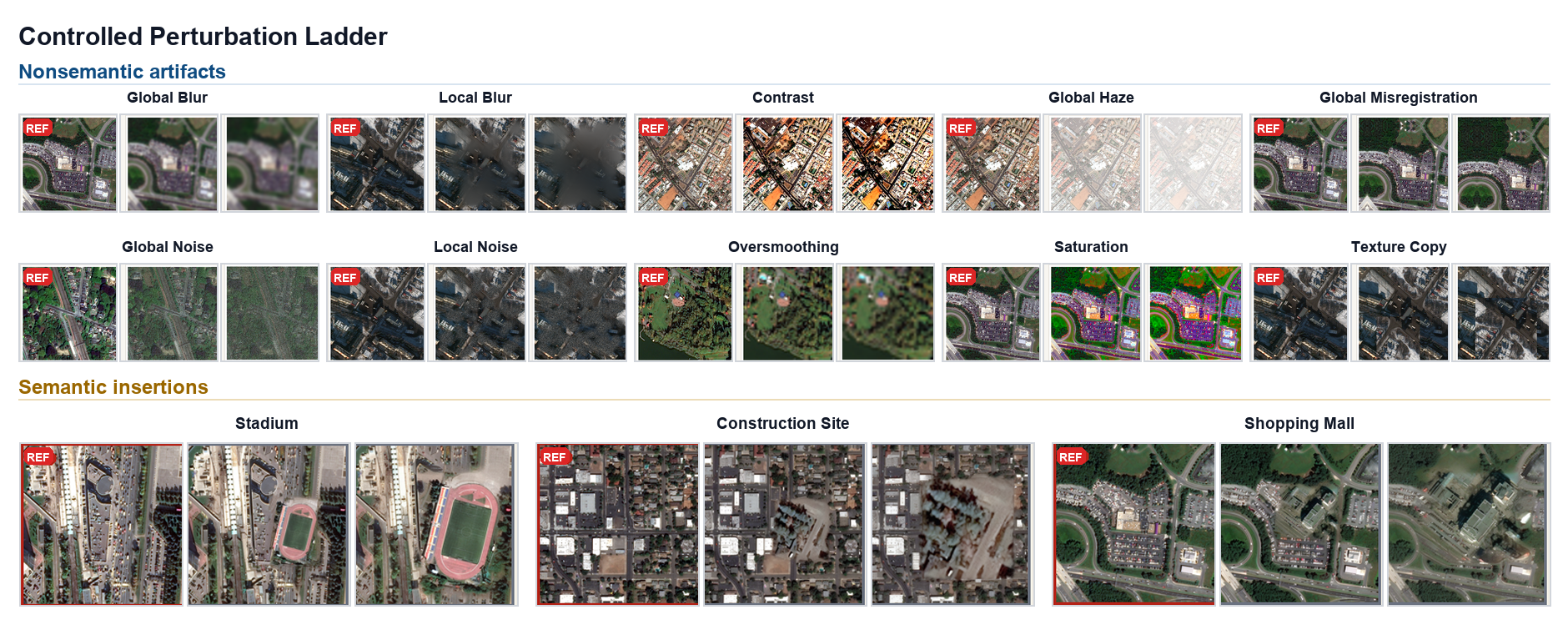}
\caption{Overview of the controlled perturbation benchmark. Background scenes are drawn from RSICD~\cite{lu2017exploring} and semantic insertions are generated with HarmoniDiff-RS~\cite{zhuang2026harmonidiffrstrainingfreediffusionharmonization}. Nonsemantic families are applied to fixed RS backgrounds with progressively stronger severities calibrated by family-specific MS-SSIM targets, while the semantic examples stage increasing object insertions under controlled area progression, yielding comparable within-family ladders without forcing different artifact types to share one pixel-space notion of severity.}
\label{fig:ladder}
\end{figure*}

\section{Controlled Synthetic Perturbations}
\label{sec:ladder}

\subsection{RemoteCLIP Aligned Attention Score}
\label{sec:method}

Our core premise is that metric design depends not only on the foundation model backbone, but also on the comparison operator. Pooling a token grid into one vector is useful for retrieval, but it can discard spatial evidence for RS failures such as scene shifts, copied roof textures, or inserted buildings. Attention offers another comparison: a query from one image evaluated against the keys of the other gives a direct statement about correspondence. We therefore measure how far the cross-image attention pattern departs from standard self-attention at each depth of RemoteCLIP~\cite{liu2024remoteclipvisionlanguagefoundation}.

Given a reference image $x$ and a candidate image $y$, we score perceptual deviation with layerwise attention agreement in RemoteCLIP instead of only comparing final pooled embeddings. For transformer block $l$, let $H_l(x)$ and $H_l(y)$ denote the hidden states entering that block. We form query, key, and value tensors
\begin{align}
Q_x, K_x, V_x &= \phi_l(H_l(x)), &
Q_y, K_y, V_y &= \phi_l(H_l(y)),
\end{align}
then compute self- and cross-attention outputs
\begin{align}
A_{xx} &= A(Q_x, K_x, V_x), & A_{xy} &= A(Q_x, K_y, V_y), \\
A_{yy} &= A(Q_y, K_y, V_y), & A_{yx} &= A(Q_y, K_x, V_x).
\end{align}
The layerwise agreement score is
\begin{equation}
\label{eq:aas}
s_l(x,y)=\tfrac{1}{2}\bigl[
\cos(A_{xy},A_{xx})+\cos(A_{yx},A_{yy})
\bigr]
\end{equation}
and the corresponding raw distance is
\begin{equation}
d^{\mathrm{raw}}_l(x,y)=\max\!\left(0,1-\mathrm{clip}(s_l(x,y),-1,1)\right).
\end{equation}
We use RemoteCLIP as the backbone because domain-aligned pretraining better captures RS-relevant semantics~\cite{liu2024remoteclipvisionlanguagefoundation}, while maintaining a lightweight, single-pass evaluation footprint.

\paragraph{Relation to Prior Work.} The single-layer operator above is adapted rather than new. Swapping keys and values between two images inside one attention layer, then scoring the agreement between the resulting cross-attention output and the corresponding self-attention output, is the Aligned Attention Score of DiffSim~\cite{song2024diffsimtamingdiffusionmodels}, defined there over the attention layers of a denoising U-Net at a selected timestep. We instead compute AAS across every block of a frozen, domain-aligned image encoder. This removes the need for noise timesteps, text conditioning, or sampling and lets us study which RemoteCLIP depths respond to specific RS artifacts (Secs.~\ref{sec:ladders}--\ref{sec:specialization}). Because the base operator is adapted from DiffSim, we benchmark against standard DiffSim in Sec.~\ref{sec:results}.

\subsection{Controlled Perturbation Ladders}
\label{sec:ladders}
Our first ablation isolates how image-comparison metrics respond to distinct classes of RS artifacts (Figure~\ref{fig:ladder}). We build the benchmark from 100 fixed high-resolution background scenes from the RSICD collection, referred to as RSIC-H~\cite{lu2017exploring}, applying one perturbation family at a time so severity can be studied without confounding interactions between artifact types. Ten nonsemantic families are included: contrast, saturation, global and local noise, global and local blur, global misregistration, oversmoothing, haze, and local texture copy. To complement these low-level degradations we add a semantic foreground insertion family generated with HarmoniDiff-RS, which introduces plausible objects while preserving scene realism~\cite{zhuang2026harmonidiffrstrainingfreediffusionharmonization}.

A central challenge is defining severity tiers that are comparable within each family. Early attempts based on fixed PSNR targets produced ladders that were numerically regular but perceptually inconsistent across artifact types, since different perturbations can induce similar pixel-wise errors while causing very different structural changes. We therefore calibrate severity with MS-SSIM~\cite{wang2003multiscale}, which better reflects multiscale structural degradation. For each nonsemantic family we estimate the approximate lower bound reachable under a family-specific search range, then define a five-tier ladder by partitioning the interval from the clean reference to that floor into equal MS-SSIM steps, yielding schedules that are monotonic, reproducible, and better aligned with visible structural corruption than pixel-domain criteria. MS-SSIM is not a meaningful severity axis for semantic insertion, so there we use a controlled five-level increasing area schedule over the HarmoniDiff-RS insertions~\cite{zhuang2026harmonidiffrstrainingfreediffusionharmonization}, increasing the inserted object footprint while constraining placement and foreground selection so that semantic severity also progresses reproducibly.

\begin{figure}[t]
\centering
\includegraphics[width=0.97\columnwidth]{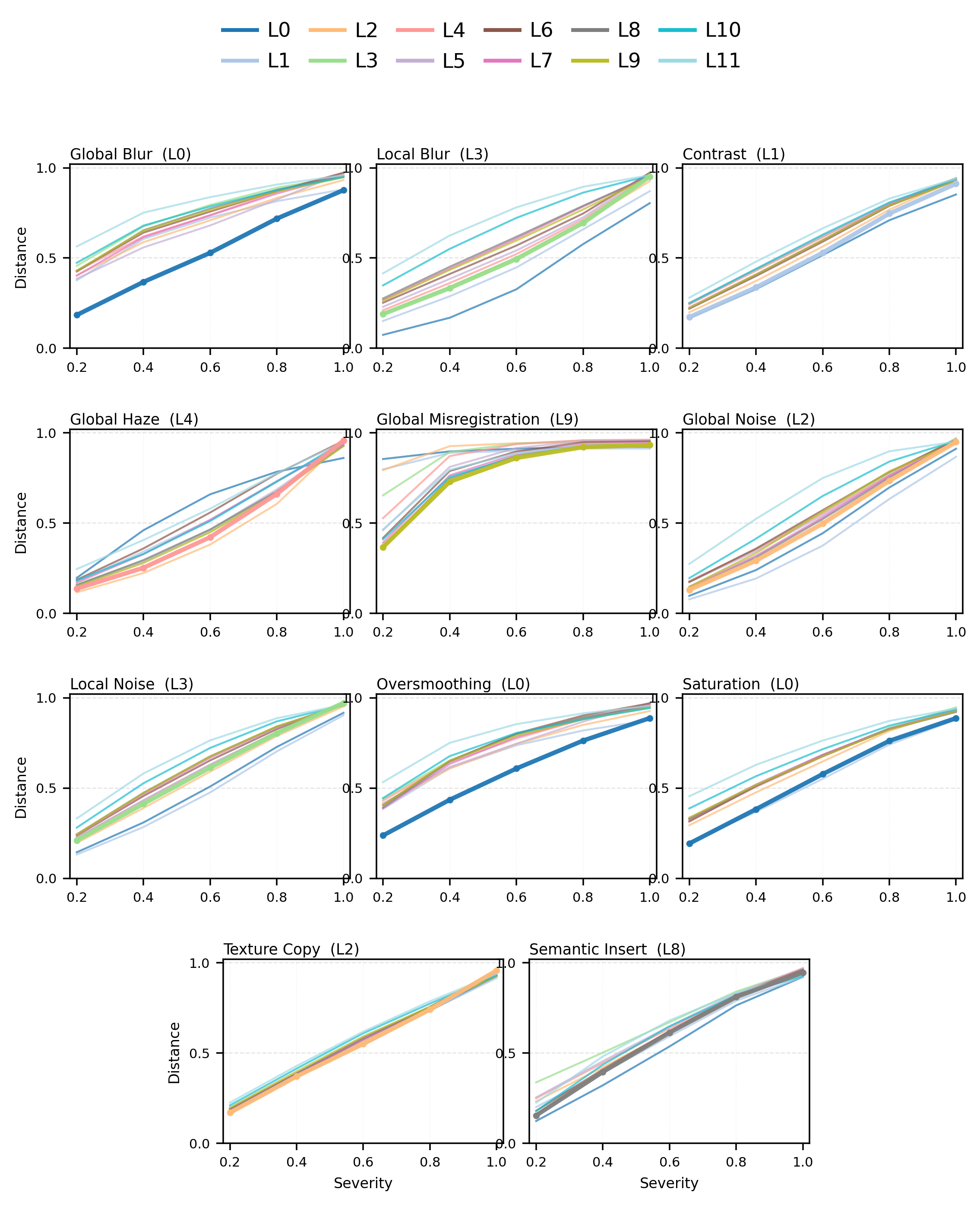}
\caption{Layerwise RemoteCLIP severity-response curves used for layer discovery. Each subpanel uses its artifact family's fixed calibration anchors; this normalization is separate from the family-independent calibration used by RSPD. The best layer changes markedly across families.}
\label{fig:all-layers}
\end{figure}

To compare layers fairly, we rescale raw distance responses using a fixed family- and layer-specific calibration anchor:
\begin{equation}
\label{eq:anchor}
a_{f,l}=\frac{1}{|S_f|}\sum_{s \in S_f}\max_{k} d^{\mathrm{raw}}_{l}(x_s, y_{f,s,k}),
\end{equation}
where $f$ indexes the perturbation family, $l$ indexes the RemoteCLIP layer, $S_f$ is the set of reference scenes for family $f$, and $y_{f,s,k}$ denotes the perturbed version of reference image $x_s$ at severity tier $k$. Calibrated distances are $d^{\mathrm{cal}}_{f,l,s,k}=\mathrm{clip}(d^{\mathrm{raw}}_{l}(x_s, y_{f,s,k})/a_{f,l},\,0,\,1)$. This is a uniform rescaling per family-layer pair, not a per-curve normalization. Crucially, it does not enforce monotonicity, widen tier separation, or otherwise reshape the response profile, but places heterogeneous layer responses on a comparable scale while preserving the structure of each severity-response curve.

Layers are then ranked by the quality of that calibrated severity-response curve, scored on four properties. We prefer responses that (i) rise from the weakest tier onward rather than only at the extreme, indicating sensitivity to mild artifacts, (ii) stay monotonic across tiers, indicating consistent ordering of severity, (iii) separate adjacent tiers rather than merging or saturating them, indicating discriminability, and (iv) span a large range between the weakest and strongest tiers, indicating that the layer uses a meaningful portion of the calibrated scale. Layers that respond only at the strongest tiers, collapse neighboring tiers, or fluctuate non-monotonically score lower.

More precisely, fix a family $f$, layer $l$, and scene $s$. For tier $i$, write the calibrated distance as $c_i=d^{\mathrm{cal}}_{f,l,s,i}$ and the change between adjacent tiers as $\delta_i=c_{i+1}-c_i$. We summarize the curve using
\begin{subequations}
\label{eq:layer-ranking}
\begin{equation}
\label{eq:curve-summaries}
\begin{aligned}
e &= \frac{4\delta_1+3\delta_2+2\delta_3+\delta_4}{10}, \\
m &= \operatorname{median}_{i}(\delta_i), \qquad r=c_5-c_1, \\
u &= \frac{1}{4}\sum_{i=1}^{4}\mathbf{1}[\delta_i\geq-\epsilon],
\end{aligned}
\end{equation}
where $e$ emphasizes early response, $m$ measures typical adjacent-tier separation, $r$ is the endpoint spread, $u$ is the fraction of nondecreasing steps, and $\epsilon=10^{-12}$. With $[z]_+=\max(0,z)$, the scene score and family-layer mean are
\begin{equation}
\label{eq:layer-ranking-score}
\begin{aligned}
\rho_{f,l,s} &= u\left(0.5[e]_++0.3[m]_++0.2[r]_+\right), \\
\bar{\rho}_{f,l} &= \frac{1}{|S_f|}\sum_{s\in S_f}\rho_{f,l,s}.
\end{aligned}
\end{equation}
\end{subequations}
We rank layers within each family by $\bar{\rho}_{f,l}$. The best layer for each family is shown in Table~\ref{tab:best-layers}; their unique set provides the seven layers used by RSPD in Sec.~\ref{sec:composite}.
\begin{table}
\centering
\small
\setlength{\tabcolsep}{5pt}
\renewcommand{\arraystretch}{0.95}
\caption{Best-performing RemoteCLIP layer for each controlled perturbation family in the controlled benchmark.}
\label{tab:best-layers}
\begin{tabular}{@{}c p{0.68\columnwidth}@{}}
\toprule
Layer & Families \\
\midrule
0 & Saturation, Global blur, Oversmoothing \\
1 & Contrast \\
2 & Global noise, Texture copy \\
3 & Local noise, Local blur \\
4 & Haze \\
8 & Semantic insertion \\
9 & Misregistration \\
\bottomrule
\end{tabular}
\end{table}

\subsection{Layer Specialization and Baseline Behavior}
\label{sec:specialization}

The ladder ablation supports two key conclusions. First, artifact sensitivity is layer-dependent (Figure~\ref{fig:all-layers}), as different families are best captured at different depths (Table~\ref{tab:best-layers}). Thus, RS artifact assessment is not well served by a single representation level. Lower and mid-level representations respond more strongly to local structure, blur, noise, and alignment error, while deeper ones react to larger-scale semantic organization. Notably, the trend is not uniform, and the best layer varies materially across families.

That variation is one argument for a composite score, but not the primary one. The ladders are deliberately idealized, varying one family at a time, whereas real generator outputs present several at once (e.g., a super-resolved tile can be simultaneously oversmoothed, slightly misregistered, and locally hallucinated, and a translated tile carries radiometric and structural error together). Because artifacts arrive aggregated, a rule that commits to the single best depth has no single family to commit to, and the depth chosen for one component of the error may be blind to another. A composite is therefore required by the structure of the failures, not only by the spread in Table~\ref{tab:best-layers}.

\begin{figure}[t]
\centering
\includegraphics[width=0.97\columnwidth]{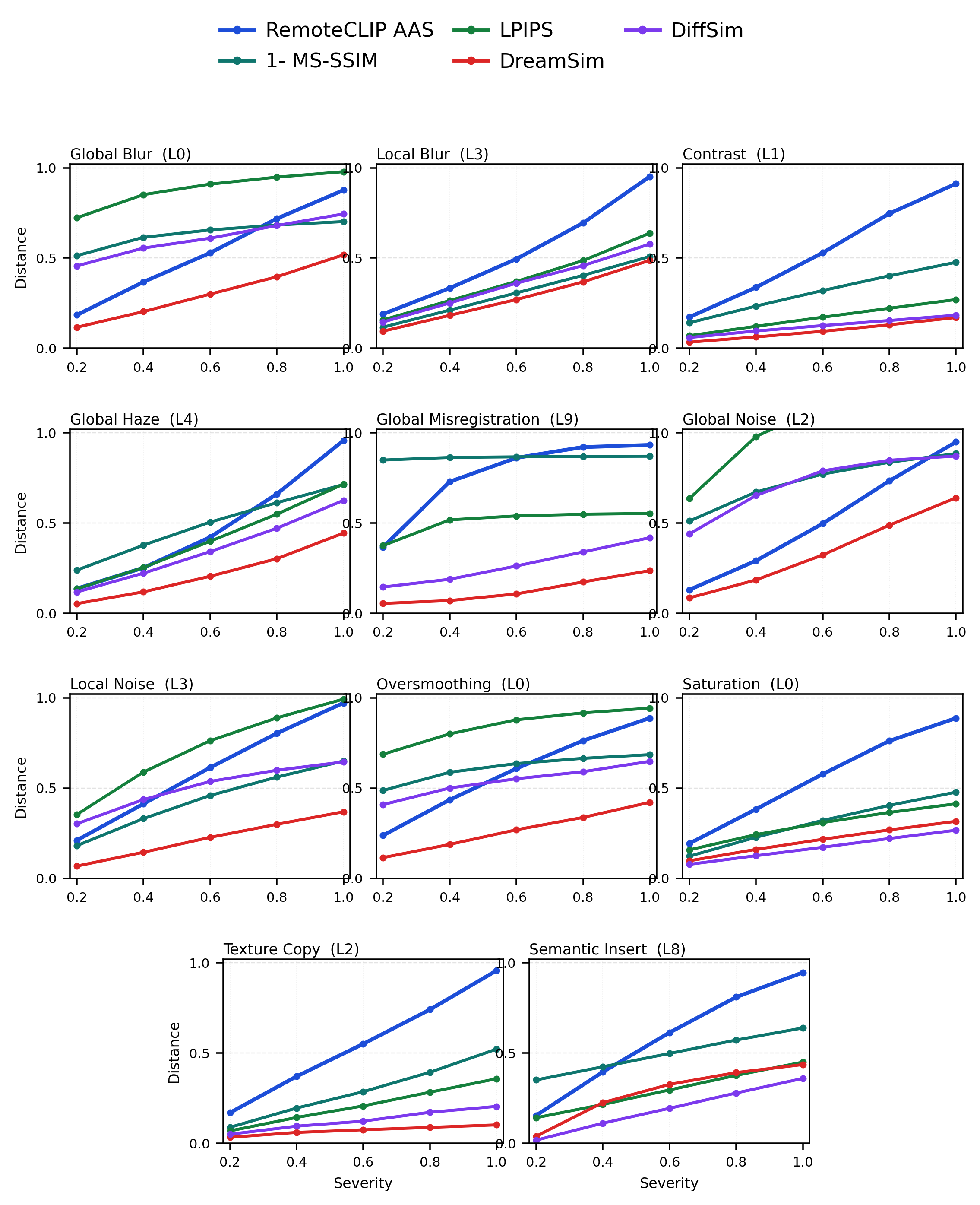}
\caption{Best family-specific RemoteCLIP \AAS{} curve against standard full-reference and learned perceptual metrics. Across many RS artifacts \AAS{} responds earlier and more monotonically, while several baselines traverse only a small part of their range or saturate at strong perturbations.}
\label{fig:metric-overlays}
\end{figure}
\begin{figure}[t]
\centering
\includegraphics[width=0.97\columnwidth]{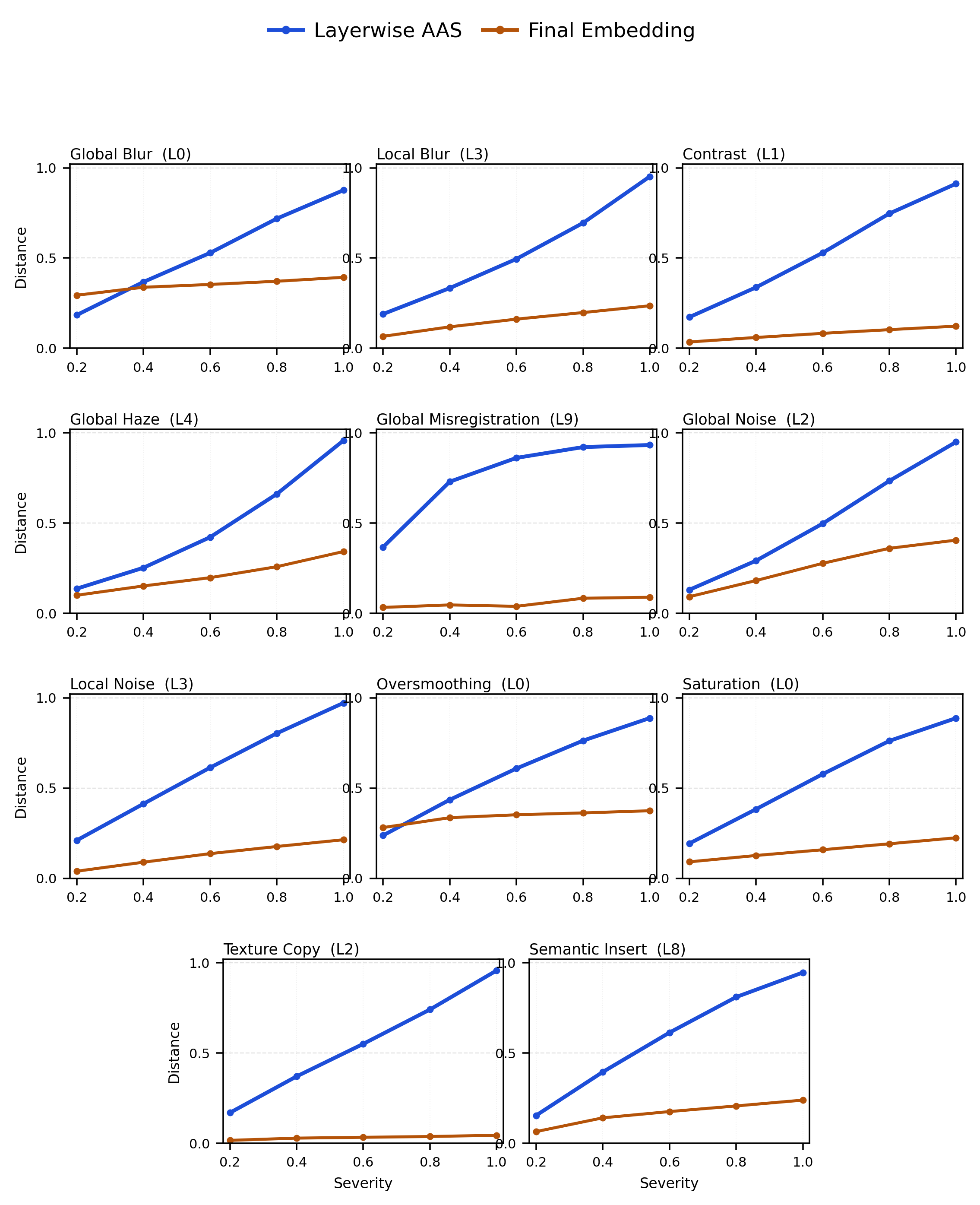}
\caption{Best family-specific \AAS{} curve versus cosine distance from the final RemoteCLIP embedding. With the checkpoint fixed, intermediate-layer AAS responds more clearly to several controlled artifacts than cosine distance on the final global embedding.}
\label{fig:embedding-vs-aas}
\end{figure}

The second conclusion concerns the baselines. Figure~\ref{fig:metric-overlays} overlays the best family-specific \AAS{} curve on MS-SSIM, LPIPS, DreamSim, and DiffSim for each family, and a couple of distinct weaknesses are visible. Regarding shape, MS-SSIM saturates or responds non-monotonically on several families, and the learned metrics stay close to flat across the mild tiers for contrast, texture copy, and semantic insertion, only reacting once the perturbation is extreme. Furthermore, magnitude emerges as a weakness, which is particularly consequential for pairwise comparison. Because the \AAS{} curves are divided by the anchor of Eq.~\eqref{eq:anchor}, their value at the strongest tier is close to $1$ by construction. The absolute endpoint is therefore not the quantity to compare, and the informative question is how much of a metric's own output range an entire severity ladder consumes. By that measure several baselines are compressed. On contrast, saturation, oversmoothing, texture copy, and semantic insertion, the learned metrics travel only a few tenths on their native scales between the clean reference and the most severe tier, encoding all five tiers within a narrow band. Such a response can be perfectly ordered and still be of little use, since a range that small is comparable to the variation the metric shows across scene content, so once two candidates each carry a mixture of degradations the ordering can invert for reasons unrelated to fidelity. This failure is distinct from being wrong on the ladder and is invisible to a metric judged only by rank correlation with severity, which is one reason a metric can look adequate on controlled ablations and still disagree with annotators on real outputs. The best \AAS{} layer for a family, by contrast, gives cleaner monotonic curves, better inter-tier separation, and more stable normalization across scenes.

\subsection{Attention-based Scoring vs. Cosine Similarity}
\label{sec:aas-vs-cosine}

The second ablation asks whether an RS-specific backbone alone suffices, or whether the scoring formulation matters too. The human-aligned RS metric of~\cite{wolters2023zooming} scores a pair by cosine similarity of two pooled CLIP embeddings, so pooled cosine is an important comparison. Using the same RemoteCLIP checkpoint, we compare the best artifact-specific \AAS{} curves against cosine distance from its final pooled embedding.

Figure~\ref{fig:embedding-vs-aas} shows the final embedding is consistently less informative than the best layerwise \AAS{} response. The gap is most pronounced for contrast, misregistration, texture copy, and semantic insertion, where final-embedding cosine stays flat across moderate tiers even though \AAS{} responds early and separates them clearly. For blur, saturation, and local noise it does vary with severity, but with weaker range and more compressed spacing, and only a few families such as global haze and global noise show broadly similar trends, with the response still attenuated.

Because the backbone is identical in both arms, this difference cannot be attributed to the RemoteCLIP checkpoint. Our comparison changes both representation depth and scoring formulation, so we make the narrower claim that intermediate-layer AAS with multi-layer aggregation is more effective here than cosine distance on the final global embedding.

\section{Pairwise Multi-Layer Scoring}
\label{sec:composite}

The metric is built in two stages: the controlled study identifies useful RemoteCLIP layers, then a frozen calibration puts their raw distances on comparable scales. The family-specific anchors in Eq.~\eqref{eq:anchor} are used only for layer discovery; they are not used when scoring a new image pair.

Let $\mathcal{L}$ denote the unique set of best family-specific layers from Table~\ref{tab:best-layers}. In our case,
\[
\mathcal{L}=\{0,1,2,3,4,8,9\}.
\]
For each layer, we pool 500 raw controlled distances from each of the 11 artifact families. Let $F_l$ be the empirical cumulative distribution function for layer $l$, linearly interpolated between observed distances and anchored at $(0,0)$. If a distance exceeds the largest calibration value $m_l$, we continue monotonically as $F_l(d)=1+(d-m_l)/q_l$, where $q_l$ is that layer's 95th-percentile calibration distance. This avoids clipping unusually severe examples.

For a reference image $x$ and candidate image $y$, RSPD is
\begin{equation}
\operatorname{RSPD}(x,y)=\frac{1}{|\mathcal{L}|}\sum_{l\in\mathcal{L}}F_l\!\left(d_l^{\mathrm{raw}}(x,y)\right).
\end{equation}
Lower scores mean greater similarity. Each reference-candidate pair receives its own score, independent of any other candidates. For a 2AFC triplet $(x,a,b)$, we simply choose $a$ when $\operatorname{RSPD}(x,a)<\operatorname{RSPD}(x,b)$ and $b$ otherwise. Neither the layer weights nor the calibration use human-preference labels. As a simpler calibration ablation, we also replace $F_l(d)$ with $d/q_l$ and refer to that variant as RSPD-Q95.

\section{Human Preference Dataset}
\label{sec:dataset}

\subsection{Triplet Design}

We cast perceptual evaluation as reference-conditioned 2AFC: each triplet presents a reference image between two candidates, and annotators select the candidate more similar to the reference. The benchmark contains five disjoint 300-triplet subsets (synthetic artifacts, airborne super-resolution, spaceborne super-resolution, SAR-to-optical, and timeseries), for 1{,}500 triplets in total.

Three design choices distinguish this collection from earlier RS preference gathering, which showed two model outputs side by side for a single task~\cite{wolters2023zooming}. First, the reference is displayed with the candidates rather than held in memory, so the task measures fidelity to a specific target rather than generic realism. This matters in RS, where the sharper, more plausible-looking output can be the less faithful one. Second, subsets from different tasks are interleaved in one blind queue, so annotators do not know which subset a triplet came from and a metric cannot be validated on a single degradation vocabulary and assumed to generalize. Third, candidates are not restricted to one artifact type. Downstream, failures arrive aggregated rather than singly, so a triplet isolating one artifact would test a capability no user needs. The synthetic subset therefore composes artifact families within a candidate, and the task-level subsets inherit their generators' mixtures.

The synthetic-artifact subset is a controlled human-preference diagnostic rather than a tuning set. The two super-resolution subsets compare paired restoration outputs against a high-resolution target, and the SAR-to-optical subset compares two translation systems against a target optical image. The timeseries subset contains curated Sentinel-2 triplets with an intentionally unambiguous ordering. It is an annotation control rather than an evaluation subset, since every metric we tested is near ceiling on it, and we use it to audit annotator attention (Sec.~\ref{sec:annotation}). Alongside these five human-annotated subsets we also evaluate on an ordered misregistration benchmark whose labels are assigned deterministically rather than by vote.

\subsection{Dataset Generation Details}

The synthetic-artifact triplets are generated from RSIC-H backgrounds, again drawn from RSICD and disjoint from the 100-reference set used in the ablations~\cite{lu2017exploring}. Unlike the controlled ladder, which isolates a single perturbation family and varies severity, the synthetic triplets contain realistic mixtures. Each candidate may include one semantic insertion and up to two nonsemantic artifact families, producing comparisons that better reflect the aggregated degradations encountered downstream.

The remaining four subsets are task-level: the two super-resolution subsets pair two model outputs against a reference target, the SAR-to-optical subset pairs two candidate translations from a SAR source against a target optical image, and the timeseries subset holds 300 curated RGB Earth-observation triplets.

Separately from the human-preference benchmark, we construct a 300-triplet ordered misregistration set in which two candidates are generated from the same reference with different magnitudes of global shift. Because larger induced shifts are less faithful to the reference by construction, we do not collect human labels here, as the candidate with smaller misregistration severity is the ground-truth preference. This gives a controlled diagnostic for a failure mode whose ordering is objectively specified.

\begin{figure}[t]
\centering
\includegraphics[width=0.95\columnwidth]{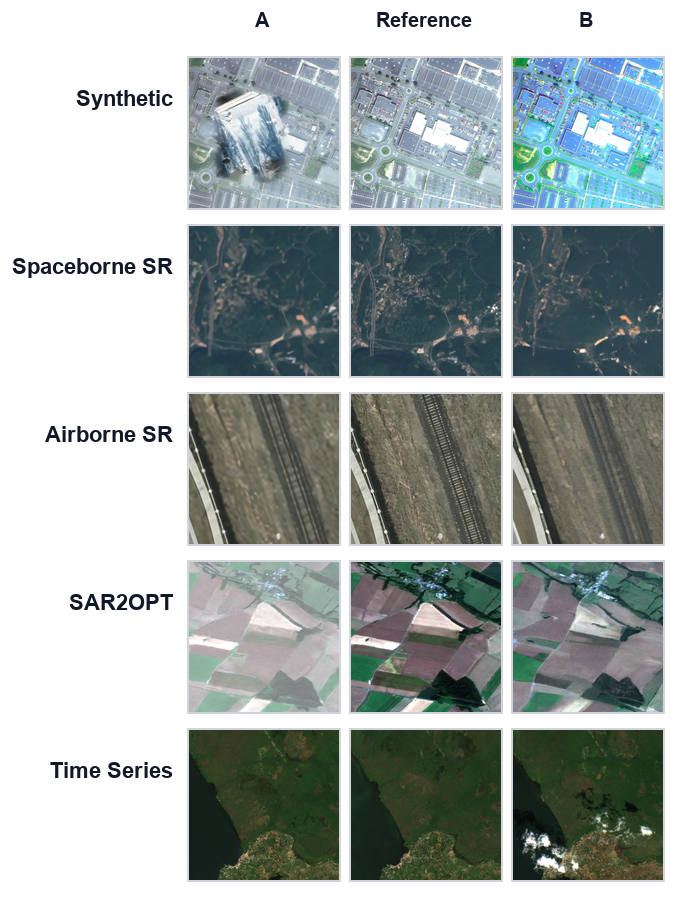}
\caption{Representative reference-conditioned 2AFC triplets in the blind annotation layout: candidate A left, reference center, candidate B right.}
\label{fig:triplet-examples}
\end{figure}

\subsection{Annotation Protocol and Label Assignment}
\label{sec:annotation}

The interface displays triplets in a blind \texttt{A|reference|B} layout, 20 per page. Annotators are instructed to choose the candidate most similar to the reference overall, considering scene structure, object layout, boundaries, textures, fine detail, and visible artifacts, and explicitly not to rely on assumptions about the underlying task, source subset, or model family. Three in-house researchers, each with remote-sensing experience, completed the annotations as part of their normal research duties, following the written instructions without additional training or screening.

To reduce order effects and annotator bias, the queue mixes triplets from all subsets, randomizes left-right placement at render time, prevents an annotator from seeing a triplet twice, and stops serving a triplet after three saved annotations. All labels come from three independent annotations per triplet, with primary labels by majority vote and unanimous subsets available for stricter analyses. Unanimous three-way agreement was $80.0\%$ for synthetic artifacts, $55.0\%$ for airborne SR, $74.7\%$ for spaceborne SR, and $84.7\%$ for SAR-to-optical.

The timeseries subset serves a different function from the rest. Because its triplets were curated to have an obvious answer, near-unanimous agreement is the expected outcome. Annotators reached unanimous three-way agreement on $92.7\%$ (278/300) of timeseries triplets, providing a consistency check on deliberately easy examples rather than establishing label reliability on the harder subsets. Every metric we evaluate is likewise near ceiling here, so the subset separates nothing and we report it as an annotation control rather than an evaluation result.

\section{Results}
\label{sec:results}

We evaluate across five subsets designed to assess both broad alignment and targeted metric behavior: three downstream human-preference subsets, a synthetic-artifact subset, and an ordered misregistration benchmark. The downstream subsets test whether RSPD, without human-preference-label tuning, competes with generic baselines on realistic RS generation. The artifact and misregistration subsets serve as controlled stress tests. In particular, the misregistration set eliminates target ambiguity because the less shifted candidate is closer by construction. (The timeseries subset is excluded as discussed in Sec.~\ref{sec:annotation}.)

All metrics are evaluated with seed 2334, deterministic PyTorch algorithms, cuDNN benchmarking disabled, and TF32 disabled. Our MS-SSIM baseline is a fixed five-scale weighted geometric mean of RGB SSIM values computed with scikit-image, using weights $(0.0448, 0.2856, 0.3001, 0.2363, 0.1333)$ and box downsampling between scales.

Alongside the natural-image baselines we report pooled-embedding cosine from the same RemoteCLIP checkpoint used by RSPD. This row is an internal comparison, not a reimplementation of the metric of~\cite{wolters2023zooming}, which pools a general web-scale CLIP encoder rather than an RS-adapted one.

The gap is large: across the five subsets of Table~\ref{tab:downstream}, pooled RemoteCLIP cosine averages $55.07$ against $89.00$ for RSPD, and on the three downstream subsets it averages $42.67$ against $90.56$. Since both use the same checkpoint, these results show that intermediate-layer AAS with multi-layer aggregation is more effective here than cosine distance on the final global embedding.

On the three downstream tasks, RSPD slightly improves over MS-SSIM on airborne SR and SAR-to-optical and matches it on spaceborne SR. MS-SSIM remains strong because these subsets pair candidates with co-registered references at matched footprints, the regime its assumptions were built for. On the misregistration set, where alignment is the variable under test, it is the weakest method in the table at $62.33$. DreamSim performs best on the synthetic and misregistration stress tests, while RSPD has the strongest overall average.

The simpler RSPD-Q95 calibration averages $87.93\%$ across the same five subsets, compared with $89.00\%$ for the ECDF calibration used by RSPD. Most of the difference comes from SAR-to-optical ($90.00\%$ versus $96.00\%$). Both variants score each candidate independently and use the same seven layers. On the timeseries control, RSPD reaches $97.00\%$ and all evaluated metrics remain near ceiling ($94.33$--$98.00\%$).

\begin{table}[t]
\centering
\caption{Reference-conditioned 2AFC accuracy (\%). The timeseries control is discussed separately. Green and yellow denote the best and second-best metrics for each dataset and the overall average.}
\label{tab:downstream}

\footnotesize
\setlength{\tabcolsep}{2.5pt}

\begin{tabular}{lcccccc}
\toprule
\textbf{Dataset} &
\textbf{MS-SSIM} &
\textbf{LPIPS} &
\shortstack{\textbf{Dream-}\\\textbf{Sim}} &
\shortstack{\textbf{Diff-}\\\textbf{Sim}} &
\shortstack{\textbf{Remote-}\\\textbf{CLIP}} &
\textbf{RSPD} \\
\midrule
Airborne SR
& \cellcolor{yellow!25}80.00
& 34.67
& 69.67
& 33.00
& 62.33
& \cellcolor{green!25}80.33 \\

Spaceborne SR
& \cellcolor{green!25}95.33
& 35.00
& \cellcolor{yellow!25}90.00
& 25.00
& 41.33
& \cellcolor{green!25}95.33 \\

SAR2Opt
& \cellcolor{yellow!25}95.00
& 62.33
& 51.00
& 69.00
& 24.33
& \cellcolor{green!25}96.00 \\

Synthetic
& 75.33
& 67.67
& \cellcolor{green!25}81.33
& 67.00
& 66.33
& \cellcolor{yellow!25}79.33 \\

Misreg.
& 62.33
& 91.67
& \cellcolor{green!25}97.33
& 88.00
& 81.00
& \cellcolor{yellow!25}94.00 \\

\midrule
\textbf{Average}
& \cellcolor{yellow!25}81.60
& 58.27
& 77.87
& 56.40
& 55.07
& \cellcolor{green!25}\textbf{89.00} \\
\bottomrule
\end{tabular}

\end{table}

\section{Discussion and Limitations}
\label{sec:discussion}

The perturbation families are non-exhaustive and only approximate the space of RS generation failures. RSPD avoids preference-label tuning but still uses a fixed seven-layer pool and calibration distribution derived from that controlled benchmark, so it may favor the artifact vocabulary represented there. We therefore treat the downstream super-resolution and SAR-to-optical subsets as the main evaluation regime and the synthetic and misregistration subsets as stress tests.

Our benchmark is also deliberately broader but smaller than the largest prior RS preference study: 1{,}500 triplets with three annotations each, against 11{,}524 pairwise annotations for super-resolution alone in~\cite{wolters2023zooming}. We trade annotation volume for task coverage, a blind mixed queue, per-subset balance, and controlled diagnostics, and regard the two collections as complementary. We cannot report agreement with that study directly, since its per-item judgments are not available to us, so our pooled-embedding comparison reimplements the operator on our own data rather than replicating theirs. Portions of the controlled benchmark use RSICD imagery that may overlap with RemoteCLIP pretraining, so the layer-selection findings should also be checked on imagery and sensors outside that distribution.

Finally, the current benchmark mainly measures RGB visual fidelity. It does not directly measure multispectral, radiometric, or SAR-domain correctness, which remain complementary parts of RS image quality.

We intend to release the benchmark, codebase, and annotations upon publication. Crucially, our framework is backbone-agnostic, requiring only accessible per-block attention. We envision this release less as a fixed leaderboard and more as a diagnostic platform to explore whether the depth-specialization patterns identified here generalize to emerging remote-sensing foundation models.

\section{Conclusion}
\label{sec:conclusion}

This paper reframes RS perceptual metric development as a dataset and benchmark problem. Prior RS work established that learned representations align with human judgment where pixel fidelity does not; we ask how those representations should be compared, and find that intermediate-layer AAS with simple multi-layer aggregation outperforms final-embedding cosine on the same RemoteCLIP checkpoint. Controlled ladders expose how RS artifacts interact with RemoteCLIP depth, while RSPD turns those findings into an independent score for any reference-candidate pair. The multi-task human-preference triplets then test whether those insights transfer downstream. The benchmark is intended to support RSPD as well as future RS metrics built on stronger backbones, alternate attention formulations, or learned aggregation.
\newpage
{
    \small
    \bibliographystyle{ieeenat_fullname}
    \bibliography{main}
}

\end{document}